%% file: main.tex
\documentclass[sigconf,nonacm,screen]{acmart}

\newcommand\blfootnote[1]{%
  \begingroup
  \renewcommand\thefootnote{}\footnote{#1}%
  \addtocounter{footnote}{-1}%
  \endgroup
}

\usepackage{graphicx}
\usepackage{booktabs}
\usepackage{textcomp}

\usepackage{url}
\usepackage[capitalise, noabbrev]{cleveref}
\usepackage{balance}
\usepackage{array,ragged2e}

\newcolumntype{R}[1]{>{\RaggedLeft\arraybackslash}p{#1}}

\newcommand{\up}{$^{\blacktriangle}$}

\newcommand{\nosign}{$^{\circ}$}

\theoremstyle{definition}

\usepackage{tikz}

\usetikzlibrary{arrows}
\usetikzlibrary{shapes}
\usepackage{setspace}

\definecolor{pastelblue}{HTML}{b3cde3}
\definecolor{pastelred}{HTML}{fbb4ae}
\definecolor{pastelpink}{HTML}{fddaec}
\definecolor{pastelgreen}{HTML}{ccebc5}
\definecolor{pastelorange}{HTML}{fed9a6}
\definecolor{pastelbrown}{HTML}{e5d8bd}

\newcommand{\boxedtex}[2]{%
  \tikz[baseline=(char.base)]
  \node[fill=#2, anchor=south west, rectangle, rounded corners, inner sep=2pt, minimum size=3mm,
        text height=2.2mm](char){#1}%
;}

\newcommand{\boxedtextsmall}[1]{%
  \tikz[baseline=(char.base)]
  \node[fill=white, anchor=south west, rectangle, rounded corners, inner sep=2pt, minimum
        size=3mm, text height=1.2mm](char){{\uppercase{\scriptsize#1}}}%
;}

\newcommand{\Annotation}[3]{\boxedtex{#1\hspace{3pt}\boxedtextsmall{#2}}{#3}}

\newenvironment{sanseriffont}{\fontfamily{phv}\selectfont\footnotesize}{\par}

\begin{document}
\input{title}
\input{01-abstract}


\maketitle
\blfootnote{Copyright \textcopyright\ 2020 for this paper by its authors. Use permitted under Creative Commons License Attribution 4.0 International (CC BY 4.0).}

\input{10-introduction}
\input{20-related-work}
\input{30-dataset}
\input{40-methods}
\input{50-results}
\input{90-conclusion}

\balance
\bibliographystyle{ACM-Reference-Format}
\bibliography{bibliography}

\end{document}

%% file: title.tex
\title{Comparing Rule-based, Feature-based and Deep Neural Methods for De-identification of Dutch Medical Records}

\author{Jan Trienes}
\affiliation{%
  \institution{Nedap Healthcare}
  \streetaddress{Parallelweg 2}
  \city{Groenlo}
  \country{Netherlands}
  \postcode{7141 DC}
}
\email{jan.trienes@nedap.com}
\orcid{0000-0001-8891-0022}

\author{Dolf Trieschnigg}
\affiliation{%
  \institution{Nedap Healthcare}
  \streetaddress{Parallelweg 2}
  \city{Groenlo}
  \country{Netherlands}
  \postcode{7141 DC}
}
\email{dolf.trieschnigg@nedap.com}

\author{Christin Seifert}
\affiliation{%
  \institution{University of Twente}
  \city{Enschede}
  \country{Netherlands}
}
\email{c.seifert@utwente.nl}

\author{Djoerd Hiemstra}
\affiliation{%
  \institution{Radboud University}
  \city{Nijmegen}
  \country{Netherlands}
}
\email{djoerd.hiemstra@ru.nl}

%% file: 01-abstract.tex
\begin{abstract}
Unstructured information in electronic health records provide an invaluable resource for medical
research.
To protect the confidentiality of patients and to conform to privacy regulations, de-identification
methods automatically remove personally identifying information from these medical records.
However, due to the unavailability of labeled data, most existing research is constrained to
English medical text and little is known about the generalizability of de-identification methods
across languages and domains.
In this study, we construct a varied dataset consisting of the medical records of 1260 patients
by sampling data from 9 institutes and three domains of Dutch healthcare.
We test the generalizability of three de-identification methods across languages and domains.
Our experiments show that an existing rule-based method specifically developed for the Dutch
language fails to generalize to this new data. Furthermore, a state-of-the-art neural architecture
performs strongly across languages and domains, even with limited training data. Compared to
feature-based and rule-based methods the neural method requires significantly less configuration
effort and domain-knowledge.
We make all code and pre-trained de-identification models available to the research
community, allowing practitioners to apply them to their datasets and to enable future benchmarks.
\end{abstract}

%% file: 10-introduction.tex
\section{Introduction}
With the strong adoption of electronic health records (EHRs), large quantities of unstructured
medical patient data become available.
This data offers significant opportunities to advance medical research and to improve healthcare
related services.
However, it has to be ensured that the privacy of a patient is protected when performing secondary
analysis of medical data.
This is not only an ethical prerequisite, but also a legal requirement imposed by privacy
legislations such as the US Health Insurance Portability and Accountability Act
(HIPAA)~\cite{US:1996:HIPAA} and the European General Data Protection Regulation
(GDPR)~\cite{EU:2016:GDPR}.
To facilitate privacy protection, de-identification has been proposed as a process that removes or
masks any kind of protected health information (PHI) of a patient such that it becomes difficult to
establish a link between an individual and the data~\cite{Meystre:2015:DUC}.
What type of information constitutes PHI is in part defined by privacy laws of the corresponding
country.
For instance, the HIPAA regulation defines 18 categories of PHI including names, geographic
locations, and phone numbers~\cite{US:2012:HIPAA}. According to the HIPAA safe-harbor rule, data is
no longer personally identifying and subject to the privacy regulation if these 18 PHI categories
have been removed.
As the GDPR does not provide such clear PHI definitions, we employ the HIPAA definitions throughout
this paper.

As most EHRs consist of unstructured, free-form text, manual de-identification is a time-consuming
and error-prone process which does not scale to the amounts of data needed for many data mining and
machine learning scenarios~\cite{Neamatullah:2008:ADF,Douglass:2004:CAD}.
Therefore, automatic de-identification methods are desirable.
Previous research proposed a wide range of methods that make use of natural language processing
techniques including rule-based matching and machine learning~\cite{Meystre:2015:DUC}.
However, most evaluations are constrained to medical records written in the English language.
The generalizability of de-identification methods across languages and domains is largely
unexplored.

To test the generalizability of existing de-identification methods, we annotated a new dataset of
1260 medical records from three sectors of Dutch healthcare: elderly care, mental care and disabled
care (\cref{sec:datasets}).
\cref{fig:example-document} shows an example record with annotated PHI.
We then compare the performance of the following three de-identification methods on this data
(\cref{sec:methods}):
\begin{enumerate}
  \setlength{\parskip}{0pt}
  \item A rule-based system named DEDUCE developed for Dutch psychiatric clinical notes~\cite{Menger:2018:DPM}
  \item A feature-based Conditional Random Field (CRF) as described in~\citet{Liu:2015:ADE}
  \item A deep neural network with a bidirectional long short-term memory architecture and a CRF output layer (BiLSTM-CRF)~\cite{Akbik:2018:CSE}
\end{enumerate}
We test the transferability of each method across three domains of Dutch healthcare. Finally, the
generalizability of the methods is compared across languages using two widely used English
benchmark corpora (\cref{sec:results}).
\input{11-example}

This paper makes three main contributions. First, our experiments show that the only openly
available de-identification method for the Dutch language fails to generalize to other
Dutch medical domains.
This highlights the importance of a thorough evaluation of the generalizability of
de-identification methods.
Second, we offer a novel comparison of several state-of-the-art de-identification methods both
across languages and domains.
Our experiments show that a popular neural architecture generalizes best even when limited amounts
of training data are available.
The neural method only considers word/character sequences which we find to be sufficient and more
robust across languages and domains compared to the structural features employed by traditional
machine learning approaches.
However, our experiments also reveal that the neural method may still experience a
substantially lower performance in new domains.
A direct consequence for de-identification practitioners is that pre-trained models require
additional fine-tuning to be fully applicable to new domains.
Third, we share our pre-trained models and code with the research community.
The creation of these resources is connected to a significant time effort and requires access to
sensitive medical data. We anticipate that this resource is of direct value to text mining
researchers.

This work was presented at the first Health Search and Data Mining Workshop (HSDM 2020)~\cite{eickhoffkimwhite}.
The implementation of the de-identification systems, pre-trained models and code for running the
experiments is available at: \url{github.com/nedap/deidentify}.

%% file: 11-example.tex
\begin{figure}
  \linespread{1.5}
  \fbox{
  \begin{minipage}{.95\columnwidth}
  \begin{sanseriffont}
    Medische overdracht Datum \Annotation{26-04-2017}{date}{pastelblue} (patiënt nr. \Annotation{64088}{id}{pastelbrown})

    Instelling \Annotation{Duinendaal}{care institute}{pastelgreen}

    Datum verrichting \Annotation{24-04-2017}{date}{pastelblue} Tijdstip 23:45

    S regel: VG ALS Heeft sonde deze is eruit, alle medicatie al gekregen. Familie is boos, dhr heeft last van slijmvorming. Is hier iets aan te doen?

    O regel: NV

    E regel: Slijmvorming

    P regel: Nu niet direct op te lossen.

    ICPC code A45.00 (Advies/observatie/voorlichting/dieet)

    Patiënt Dhr. \Annotation{Jan P. Jansen}{name}{pastelorange} (M), \Annotation{06-11-1956}{date}{pastelblue} Arts \Annotation{J.O. Besteman}{name}{pastelorange} Adres \Annotation{Wite Mar 782 Kamerik}{address}{pastelpink}

    Verrichting Telefonisch consult ANW
    (t: \Annotation{06-7802651}{phone/fax}{pastelred})

    ==================== English Translation ====================

    Medical transfer date \Annotation{26-04-2017}{date}{pastelblue} (patient no. \Annotation{64088}{id}{pastelbrown})

    Institution \Annotation{Duinendaal}{care institute}{pastelgreen}

    Date \Annotation{24-04-2017}{date}{pastelblue} Time 23:45

    Subjective (S): VG ALS got feeding tube removed, already received all medication. Family is upset, Mr. suffers from increased mucus formation. Can anything be done about that?

    Objective (O): NV

    Evaluation (E): Mucus formation

    Plan (P): Cannot be solved immediately.

    ICPC code A45.00 (Advice/observation/information/diet)

    Patient Mr. \Annotation{Jan P. Jansen}{name}{pastelorange} (M), \Annotation{06-11-1956}{date}{pastelblue} Doctor \Annotation{J.O. Besteman}{name}{pastelorange} Address \Annotation{Wite Mar 782 Kamerik}{address}{pastelpink}

    Provided phone consult ANW
    (t: \Annotation{06-7802651}{phone/fax}{pastelred})

  \end{sanseriffont}
  \end{minipage}
  }
  \caption{Excerpt of a medical record in our dataset with annotated protected health information (PHI). Sensitive PHI was replaced with automatically generated surrogates.}
  \label{fig:example-document}
\end{figure}

%% file: 20-related-work.tex
\section{Related Work}\label{sec:related-work}
Previous work on de-identification can be roughly organized into four groups: (1) creation of
benchmark corpora, (2) approaches to de-identification, (3) work on languages other than English,
and (4) cross-domain de-identification.

Various English benchmark corpora have been created including nursing notes, longitudinal patient
records and psychiatric intake notes~\cite{Neamatullah:2008:ADF,Stubbs:2015:ALC,Stubbs:2017:DPI}.
Furthermore,~\citet{Deleger:2012:BGS} created a heterogeneous dataset comprised of 22 different
document types.
Contrary to the existing datasets which only contain records from at most two different medical
institutes, the data used in this paper was sampled from a total of 9 institutes that are active
in the Dutch healthcare sector. The contents, structure and writing style of the documents strongly
depend on the processes and individuals specific to an institute which contributes to a
heterogeneous corpus.

Most existing de-identification approaches are either rule-based or machine learning based.
Rule-based methods combine various heuristics in form of patterns, lookup lists and fuzzy string
matching to identify PHI~\cite{Gupta:2004:EDS,Neamatullah:2008:ADF}.
The majority of machine learning approaches employ feature-based CRFs~\cite{Aberdeen:2010:MIS,He:2015:CBD},
ensembles combining CRFs with rules~\cite{Stubbs:2015:ASD} and most recently also neural networks
\cite{Dernoncourt:2017:DPN,Liu:2017:DCN}.
A thorough overview of the different de-identification methods is given in~\citet{Meystre:2015:DUC}.
In this study, we compare several state-of-the-art de-identification methods.
With respect to rule-based approaches, we apply DEDUCE, a recently developed method for Dutch data~\cite{Menger:2018:DPM}.
To the best of our knowledge, this is the only openly available de-identification method
tailored to Dutch data.
For a feature-based machine learning method, we re-implement the token-level CRF by~\citet{Liu:2015:ADE}.
Previous work on neural de-identification used a BiLSTM-CRF architecture with character-level and
ELMo embeddings~\cite{Dernoncourt:2017:DPN,Khin:2018:DLA}. Similarly, we use a BiLSTM-CRF but apply
recent advances in neural sequence modeling by using contextual string embeddings~\cite{Akbik:2018:CSE}.

To the best of our knowledge, we are the first study to offer a comparison of de-identification
methods across languages. With respect to de-identification in languages other than English, only
three studies consider Dutch data.~\citet{Scheurwegs:2013:DCF} applied a Support Vector Machine and
a Random Forest classifier to a dataset of 200 clinical records.~\citet{Menger:2018:DPM} developed
and released a rule-based method on 400 psychiatric nursing notes and treatment plans of a single
Dutch hospital.~\citet{TjongKimSang:2019:DDM} evaluated an existing named entity tagger for the
de-identification of autobiographic emails on publicly available Wikipedia texts.
Furthermore, de-identification in several other languages has been studied including German,
French, Korean and Swedish~\cite{Richter:2018:DGM,Neveol:2018:CNL}.

With respect to cross-domain de-identification, the 2016 CEGS N-GRID shared task evaluated the
portability of pre-trained de-identification methods to a new set of English psychiatric
records~\cite{Stubbs:2017:DPI}.
Overall, the existing systems did not perform well on the new data. Here, we provide a similar
comparison by cross-testing on three domains of Dutch healthcare.

%% file: 30-dataset.tex
\section{Datasets}\label{sec:datasets}
This section describes the construction of our Dutch benchmark dataset called NUT (Nedap/University of Twente). The data was sampled from
9 healthcare institutes and annotated for PHI according to a tagging scheme derived
from~\citet{Stubbs:2015:ALC}. Furthermore, following common practice in the preparation of
de-identification corpora, we replaced PHI instances with realistic surrogates to comply with
privacy regulations. To compare the performance of the de-identification methods across languages,
we use the English i2b2/UTHealth and the nursing notes corpus~\cite{Stubbs:2015:ALC,Neamatullah:2008:ADF}.
An overview of the three datasets can be found in~\cref{tab:dataset-statistics}.
\begin{table}[t]
\caption{Overview of the datasets used in this study.}
\label{tab:dataset-statistics}
\begin{tabular}{@{}lR{.27\columnwidth}rr@{}}
\toprule
Datset & NUT & i2b2~\cite{Stubbs:2015:ALC} & Nursing~\cite{Neamatullah:2008:ADF} \\ \midrule
Language & Dutch & English & English \\
Domain(s) & elderly, mental and disabled care & clinical & clinical \\
Institutes & 9 (3 per domain) & 2 & 1 \\
Documents & 1260 & 1304 & 2434 \\
Patients & 1260 & 296 & 148 \\
Tokens & 448,795 & 1,057,302 & 444,484 \\
Vocabulary & 25,429 & 36,743 & 19,482 \\
PHI categories & 16 & 32 & 10 \\
PHI instances & 17,464 & 28,872 & 1779 \\
Median PHI/doc. & 9 & 18 & 0 \\ \bottomrule
\end{tabular}%
\end{table}
\subsection{Data Sampling}
We sample data from a snapshot of the databases of 9 healthcare institutes with a total of 83,000
patients.
Three domains of healthcare are equally represented in this snapshot: elderly care, mental care and
disabled care.
We consider two classes of documents to sample from: surveys and progress reports.
Surveys are questionnaire-like forms which are used by the medical staff to take notes during
intake interviews, record the outcomes of medical tests or to formalize the treatment plan of a
patient.
Progress reports are short documents describing the current conditions of a patient receiving care,
sometimes on a daily basis.
The use of surveys and progress reports differs strongly across healthcare institute and domain.
In total, this snapshot consists of 630,000 surveys and 13 million progress reports.

When sampling from the snapshot described above, we aim to maximize both the variety of document
types, and the variety of PHI, two essential properties of a de-identification benchmark corpus
\cite{Deleger:2012:BGS}.
First, to ensure a wide variety of document types, we select surveys in a stratified fashion
according to their type label provided by the EHR system (e.g., intake interview, care plan, etc.).
Second, to maximize the variety in PHI, we sample medical reports on a patient basis: for each
patient, a random selection of 10 medical reports is combined into a patient file.
We then select patient files uniformly at random to ensure that no patient appears multiple
times within the sample.
Furthermore, to control the annotation effort, we impose two subjective limits on the document
length.
A document has to contain at least 50 tokens, but no more than 1000 tokens to be included in the
sample.
For each of the 9 healthcare institutes, we sample 140 documents (70 surveys and 70 patient files),
which yields a total sample size of 1260 documents (see~\cref{tab:dataset-statistics}).

We received approval for the collection and use of our dataset from the ethics review board of our institution. Due to privacy regulations, the dataset constructed in this paper cannot be shared.

\subsection{Annotation Scheme}
Since the GDPR does not provide any strict rules about which types of PHI should be removed during
de-identification, we base our PHI tagging scheme on the guidelines defined by the US HIPAA
regulations.
In particular, we closely follow the annotation guidelines and the tagging scheme used by
\citet{Stubbs:2015:ALC} which consists of 32 PHI tags among 8 classes: \emph{Name},
\emph{Profession}, \emph{Location}, \emph{Age}, \emph{Date}, \emph{Contact Information}, \emph{IDs}
and \emph{Other}.
The \emph{Other} category is used for information that can be used to identify a patient, but which
does not fall into any of the remaining categories.
For example, the sentence \emph{``the patient was a guest speaker on the subject of diabetes in the
Channel 2 talkshow.''} would be tagged as \emph{Other}.
It is worth mentioning that this tagging scheme does not only capture direct identifiers relating
to a patient (e.g., name and date of birth), but also indirect identifiers that could be used in
combination with other information to reveal the identity of a patient.
Indirect identifiers include, for example, the doctor's name, information about the
hospital and a patient's profession.

\begin{table}[t]
\caption{PHI tags used to annotate our dataset (NUT). The tagging scheme was derived from the i2b2 tags.}
\label{tab:annotation-scheme}
\begin{tabular}{@{}lp{.4\columnwidth}l@{}}
\toprule
Category & i2b2~\cite{Stubbs:2015:ALC} & NUT \\ \midrule
Name & Patient, Doctor, Username & Name \\
 &  & Initials \\
Profession & Profession & Profession \\
Location & Room, Department & Internal Location \\
 & Hospital, Organization & Hospital, Organization \\
 &  & Care Institute \\
 & Street, City, State, ZIP, Country & Address \\
Age & Over 90, Under 90 & Age \\
Date & Date & Date \\
Contact & Phone, FAX, Email & Phone/FAX, Email \\
 & URL, IP & URL/IP \\
IDs & SSN, 8 fine-grained ID tags & SSN, ID \\
Other & Other & Other \\ \bottomrule
\end{tabular}%
\end{table}

We made two adjustments to the tagging scheme by~\citet{Stubbs:2015:ALC}.
First, to reduce the annotation effort, we merged some of the 32 fine-grained PHI tags to a more
generic set of 16 tags (see~\cref{tab:annotation-scheme}).
For example, the fine-grained location tags \emph{Street}, \emph{City}, \emph{State}, \emph{ZIP},
and \emph{Country} were merged into a generic \emph{Address} tag.
While this simplifies the annotation process, it complicates the generation of realistic surrogates.
Given an address string, one has to infer its format to replace the individual parts with
surrogates of the same semantic type.
We address this issue in~\cref{sec:surrogate-generation}.
Second, due to the high frequency of care institutes in our dataset, we decided to introduce a
separate \emph{Care Institute} tag that complements the \emph{Organization} tag.
This allows for a straightforward surrogate generation where names of care institute are replaced
with another care institute rather than with more generic company names (e.g., Google).

\subsection{Annotation Process}
Following previous work on the construction of de-identification benchmark corpora
\cite{Stubbs:2015:ALC,Deleger:2012:BGS}, we employ a double-annotation strategy: two annotators
read and tag the same documents.
In total, 12 non-domain experts annotated the sample of 1260 medical records independently
and in parallel.
The documents were randomly split into 6 sets and we randomly assigned a pair of annotators to each set.
To ensure that the annotators had a common understanding of the annotation instructions, an
evaluation session was held after each pair of annotators completed the first 20 documents.%
\footnote{
We include the annotation instructions that were provided to the annotators in the online repository
of this paper.
The instructions are in large parts based on the annotation guidelines in~\citet{Stubbs:2015:ALC}.
}
In total, it took 77 hours to double-annotate the entire dataset of 1260 documents, or
approximately 3.7 minutes per document.
We measured the inter-annotator agreement (IAA) using entity-level F1 scores.%
\footnote{%
  It has been shown that the F-score is more suitable to quantify IAA in sequence-tagging scenarios
  compared to other measures such as the Kappa score~\cite{Deleger:2012:BGS}.
}
\cref{tab:phi-distribution} shows the IAA per PHI category. Overall, the agreement level is
fairly high (0.84). However, we find that location names (i.e., care institutes, hospitals,
organizations and internal locations) are often highly ambiguous which is reflected by the low
agreement scores of these categories (between 0.29 and 0.52).

\begin{table}[t]
\caption{Distribution of PHI tags in our dataset. The inter-annotator agreement (IAA) as
measured by the micro-averaged F1 score is shown per category.}
\label{tab:phi-distribution}
\begin{tabular}{@{}lrrr@{}}
\toprule
PHI Tag & Count & Frac. (\%) & IAA\\ \midrule
Name & 9558 & 54.73 & 0.96\\
Date & 3676 & 21.05 & 0.86 \\
Care Institute & 997 & 5.71 & 0.52\\
Initials & 778 & 4.45 & 0.46 \\
Address & 748 & 4.28 & 0.75 \\
Organization & 712 & 4.08 & 0.38 \\
Internal Location & 242 & 1.39 & 0.29 \\
Age & 175 & 1.00 & 0.39\\
Profession & 122 & 0.70 & 0.31 \\
ID & 114 & 0.65 & 0.43 \\
Phone/Fax & 97 & 0.56 & 0.93 \\
Email & 95 & 0.54 & 0.94 \\
Hospital & 92 & 0.53 & 0.42 \\
Other & 33 & 0.19 & 0.03 \\
URL/IP & 23 & 0.13 & 0.70 \\
SSN & 2 & 0.01 & 0.50 \\ \midrule
Total & 17,464 & 100 & 0.84 \\ \bottomrule
\end{tabular}%
\end{table}

To improve annotation efficiency, we integrated the rule-based de-identification tool
DEDUCE~\cite{Menger:2018:DPM} with our annotation software to pre-annotate each document.
This functionality could be activated on a document basis by each annotator.
If an annotator used this functionality, they had to review the pre-annotations, correct potential
errors and check for missed PHI instances.
During the evaluation sessions, annotators mentioned that the existing tool proved helpful when
annotating repetitive names, dates and email addresses.
Note that this pre-annotation strategy might give DEDUCE a slight advantage. However, the low
performance of DEDUCE in the formal benchmark in~\cref{sec:results} does not reflect this.\looseness=-1

After annotation, the main author of this paper reviewed 19,165 annotations and resolved any
disagreements between the two annotators to form the gold-standard of 17,464 PHI annotations.
\cref{tab:phi-distribution} shows the distribution of PHI tags after adjudication.
Overall the adjudication has been done risk-averse: if only one annotator identified a piece of
text as PHI, we assume that the other annotator missed this potential PHI instance.
In addition to the manual adjudication, we performed two automatic checks: (1) we ensured that PHI
instances occurring in multiple files received the same PHI tag, and (2) any instances that were
tagged in one part of the corpora but not in the other were manually reviewed and added to the
gold-standard.
We used the BRAT annotation tool for both annotation and adjudication~\cite{Stenetorp:2012:BWT}.

\subsection{Surrogate Generation}\label{sec:surrogate-generation}
As the annotated dataset consists of personally identifying information which is protected by the
GDPR, we generate artificial replacements for each of the PHI instances before using the data for
the development of de-identification methods.
This process is known as surrogate generation, a common practice in the preparation of
de-identification corpora~\cite{Stubbs:2015:CSS}.
As surrogate generation will inevitably alter the semantics of the corpus to an extent where it
affects the de-identification performance, it is important that this step is done as thoroughly as
possible~\cite{Yeniterzi:2010:EPI}.
Here, we follow the semi-automatic surrogate generation procedure that has been used to prepare the
i2b2/UTHealth shared task corpora.
Below, we summarize this procedure and mention the language specific resources we used.
We refer the reader to~\citet{Stubbs:2015:CSS} for a thorough discussion of the method.
After running the automatic replacement scripts, we reviewed each of the surrogates to ensure that
continuity within a document is preserved and no PHI is leaked into the new dataset.

We adapt the surrogate generation method of~\citet{Stubbs:2015:CSS} to the Dutch language as
follows. A list of 10,000 most common family names and given names is used to generate random
surrogates for name PHI instances.\footnote{See~\url{www.naamkunde.net}, accessed 2019-12-09}
We replace dates by first parsing the format (e.g., ``12 nov. 2018'' $\rightarrow$ ``\%d \%b. \%Y''),\footnote{Rule-based date parser:~\url{github.com/nedap/dateinfer}, accessed 2019-12-09} and then
randomly shifting all dates within a document by the same amount of years and days into the future.
For addresses, we match names of cities, streets, and countries with a dictionary of Dutch
locations,\footnote{See~\url{openov.nl}, accessed 2019-12-09} and then pick random replacements from that
dictionary.
As Dutch ZIP codes follow a standard format (``1234AB''), their replacement is straightforward.
Names of hospitals, care institutes, organizations and internal locations are randomly shuffled
within the dataset.
PHI instances of type \emph{Age} are capped at 89 years. Finally, alphanumeric strings such as
\emph{Phone/FAX}, \emph{Email}, \emph{URL/IP}, \emph{SSN} and \emph{IDs} are replaced by
substituting each alphanumeric character with another character of the same class.
We manually rewrite \emph{Profession} and \emph{Other} tags, as an automatic replacement is not
applicable.

%% file: 40-methods.tex
\section{Methods}\label{sec:methods}
This section presents the three de-identification methods and the evaluation procedure.

\subsection{Rule-based Method: DEDUCE}
DEDUCE is an unsupervised de-identification method specifically developed for Dutch medical records
\cite{Menger:2018:DPM}.
It is based on lookup tables, decision rules and fuzzy string matching and has been validated
on a corpus of 400 psychiatric nursing notes and treatment plans of a single hospital.
Following the authors' recommendations, we customize the method to include a list of 1200
institutions that are common in our domain.
Also, we resolve two incompatibilities between the PHI coding schemes of our dataset and the DEDUCE
output.
First, as DEDUCE does not distinguish between hospitals, care institutes, organizations and
internal locations, we group these four PHI tags under a single \emph{Named Location} tag.
Second, our \emph{Name} annotations do not include titles (e.g., ``Dr.'' or ``Ms.''). Therefore,
titles are stripped from the DEDUCE output.

\subsection{Feature-based Method: Conditional Random Field}
CRFs and hybrid rule-based systems provide state-of-the-art performance
in recent shared tasks~\cite{Stubbs:2015:ASD,Stubbs:2017:DPI}.
Therefore, we implement a CRF approach to contrast with the unsupervised rule-based system.
In particular, we re-implement the token-based CRF method by~\citet{Liu:2015:ADE} and re-use a subset\footnote{We disregard word-representation features as~\citet{Liu:2015:ADE} found that they had a negative performance impact.} of their features (see~\cref{tab:crf-features}).
The linear-chain CRF is trained using LBFGS and elastic net regularization~\cite{Zou:2005:RVS}.
Using a validation set, we optimize the two regularization coefficients of the $L_1$ and $L_2$
norms with a random search in the $log_{10}$ space of $[10^{-4},10^1]$ with 250 trials.
We use the \emph{CRFSuite} implementation by~\citet{Okazaki:2007:CRFsuite}.

\begin{table}[t]
\caption{Features used by the CRF method. The features are identical to the one by~\citet{Liu:2015:ADE}, but we exclude word-representation features.}
\label{tab:crf-features}
\begin{tabular}{@{}p{.35\columnwidth}p{.58\columnwidth}@{}}
\toprule
Group & Description \\ \midrule
Bag-of-words (BOW) & Token unigrams, bigrams and trigrams within a window of $[-2,2]$ of the current token. \\
Part-of-speech (POS) & Same as above but with POS n-grams. \\
BOW + POS & Combinations of the previous, current and next token and their POS tags. \\
Sentence & Length in tokens, presence of end-mark such as '.', '?', '!' and whether sentence contains unmatched brackets. \\
Affixes & Prefix and suffix of length 1 to 5. \\
Orthographic & Binary indicators about word shape: is all caps, is capitalized, capital letters inside, contains digit, contains punctuation, consists of only ASCII characters. \\
Word Shapes & The abstract shape of a token. For example, ``7534-Df'' becomes ``\#\#\#\#-Aa''.\\
Named-entity recognition (NER) & NER tag assigned by the spaCy tagger.\\
\bottomrule
\end{tabular}%
\end{table}

\subsection{Neural Method: BiLSTM-CRF}
To reduce the need for hand-crafted features in traditional CRF-based de-identification,
recent work applies neural methods~\cite{Liu:2017:DCN,Dernoncourt:2017:DPN,Khin:2018:DLA}.
Here, we re-implement a BiLSTM-CRF architecture with contextual string embeddings, which has
recently shown to provide state-of-the-art results for sequence labeling
tasks~\cite{Akbik:2018:CSE}.
Hyperparameters are set to the best performing configuration in~\citet{Akbik:2018:CSE}:
we use stochastic gradient descent with no momentum and an initial learning rate of 0.1. If the
training loss does not decrease for 3 consecutive epochs, the learning rate is halved.
Training is stopped if the learning rate falls below $10^{-4}$ or 150 epochs are reached.
Furthermore, the number of hidden layers in the LSTM is set to 1 with 256 recurrent units.
We employ locked dropout with a value of 0.5 and use a mini-batch size of 32.
With respect to the embedding layer, we use the pre-trained GloVe (English) and fasttext (Dutch)
embedding on a word-level, and concatenate them with the pre-trained contextualized string
embeddings included in Flair\footnote{\url{github.com/zalandoresearch/flair}, accessed 2019-12-09}~\cite{Pennington:2014:GGV,Grave:2018:LWV,Akbik:2019:PCE}.

\subsection{Preprocessing and Sequence Tagging}
We use a common preprocessing routine for all three datasets. For tokenization and sentence
segmentation, the spaCy tokenizer is used.\footnote{\url{spacy.io}, accessed 2019-12-09}
The POS/NER features of the CRF method are generated by the built-in spaCy models.
After sentence segmentation, we tag each token according to the Beginning, Inside, Outside
(BIO) scheme.
In rare occasions, sequence labeling methods may produce invalid transitions (e.g., {\tt O-} $\rightarrow$ {\tt I-}). In a post-processing step, we replace invalid {\tt I-} tags with {\tt B-} tags~\cite{Reimers:2017:OHD}.

\subsection{Evaluation}\label{sec:evaluation}
The de-identification methods are assessed according to precision, recall and F1 computed on an entity-level, the standard evaluation approach for NER systems~\cite{TjongKimSang:2003:ICS}.
In an entity-level evaluation, predicted PHI offsets and types have to match exactly.
Following the evaluation of de-identification shared tasks, we use the micro-averaged entity-level F1 score as primary metric~\cite{Stubbs:2015:ASD}.\footnote{%
De-identification systems are often also evaluated on a less strict token-level. As a system
that scores high on an entity-level will also score high on a token-level, we only
measure according to the stricter level of evaluation.
}

We randomly split our dataset and the nursing notes corpus into training, validation and testing sets with a 60/20/20 ratio.
As the i2b2 corpus has a pre-defined test set of 40\%, a random set of 20\% of the training
documents serves as validation data.
Finally, we test for statistical significance using two-sided approximate randomization with $N=9999$~\cite{Yeh:2000:MAT}.

%% file: 50-results.tex
\begin{table*}[t]
\caption{%
Evaluation summary: micro-averaged scores are shown for each dataset and method.
Statistically significant improvements over the score on the previous line are marked with \up ($p<0.01$), and \nosign\ depicts no significance. The rule-based method DEDUCE is not applicable to the English datasets.
}
\label{tab:results-summary}
\begin{tabular}{@{}llllllllll@{}}
\toprule
 & \multicolumn{3}{c}{NUT (Dutch)} & \multicolumn{3}{c}{i2b2 (English)} & \multicolumn{3}{c}{Nursing Notes (English)} \\
 \cmidrule(lr){2-4}
 \cmidrule(lr){5-7}
 \cmidrule(lr){8-10}
Method & Prec. & Rec. & F1 & Prec. & Rec. & F1 & Prec. & Rec. & F1 \\ \midrule
DEDUCE & 0.807 & 0.564 & 0.664 & - & - & - & - & - & - \\
CRF & \textbf{0.919}\up & 0.775\up & 0.841\up & 0.952 & 0.796 & 0.867 & \textbf{0.914} & 0.685 & 0.783 \\
BiLSTM-CRF &
 0.917\nosign & \textbf{0.871}\up & \textbf{0.893}\up &
 \textbf{0.959}\up & \textbf{0.869}\up & \textbf{0.912}\up &
 0.886\nosign & \textbf{0.797}\up & \textbf{0.839}\up \\
\bottomrule
\end{tabular}
\end{table*}

\section{Results}\label{sec:results}
In this section, we first discuss the de-identification results obtained on our Dutch dataset (\cref{sec:results-dutch-data}).
Afterwards, we present an error analysis of the best performing method (\cref{sec:error-analysis}).
This section is concluded with the benchmark for the English datasets (\cref{sec:results-english-data}) and the cross-domain de-identification (\cref{sec:results-cross-domain}).

\begin{table}[t]
\caption{Entity-level precision and recall per PHI category on the NUT dataset. Scores are
compared between the rule-based tagger DEDUCE~\cite{Menger:2018:DPM} and the BiLSTM-CRF model. The
\textit{Named Loc.} tag is the union of the 4 specific location tags which are not supported by
DEDUCE. Tags are ordered by frequency with location tags fixated at the bottom.}
\label{tab:phi-breakdown}
\begin{tabular}{lrrrr}
\toprule
 & \multicolumn{2}{c}{BiLSTM-CRF} & \multicolumn{2}{c}{DEDUCE}  \\
\cmidrule(lr){2-3}
\cmidrule(lr){4-5}
PHI Tag & Prec. & Rec. & Prec. & Rec. \\ \midrule
Name & \textbf{0.965} & \textbf{0.956} & 0.849 & 0.805 \\
Date & \textbf{0.926} & \textbf{0.920} & 0.857 & 0.441 \\
Initials & \textbf{0.828} & \textbf{0.624} & 0.000 & 0.000 \\
Address & \textbf{0.835} & \textbf{0.846} & 0.804 & 0.526 \\
Age & \textbf{0.789} & \textbf{0.732} & 0.088 & 0.122 \\
Profession & \textbf{0.917} & \textbf{0.262} & 0.000 & 0.000 \\
ID & \textbf{0.800} & \textbf{0.480} & 0.000 & 0.000 \\
Phone/Fax & 0.889 & \textbf{1.000} & \textbf{0.929} & 0.812 \\
Email & 0.909 & \textbf{1.000} & \textbf{1.000} & 0.900 \\
Other & 0.000 & 0.000 & 0.000 & 0.000 \\
URL/IP & \textbf{1.000} & \textbf{0.750} & 0.750 & \textbf{0.750} \\
Named Loc. & \textbf{0.797} & \textbf{0.659} & 0.279 & 0.058 \\ \midrule
\hspace{3mm}Care Institute & 0.686 & 0.657 & n/a & n/a \\
\hspace{3mm}Organization & 0.780 & 0.522 & n/a & n/a \\
\hspace{3mm}Internal Loc. & 0.737 & 0.509 & n/a & n/a \\
\hspace{3mm}Hospital & 0.778 & 0.700 & n/a & n/a \\
\bottomrule
\end{tabular}
\end{table}

\subsection{De-identification of Dutch Dataset}\label{sec:results-dutch-data}
Both machine learning methods outperform the rule-based system DEDUCE by a large margin (see~\cref{tab:results-summary}).
Furthermore, the BiLSTM-CRF provides a substantial improvement of 10\% points in recall over the
traditional CRF method, while maintaining precision.
Overall, the neural method has an entity-level recall of 87.1\% while achieving a recall of 95.6\%
for names, showing that the neural method is operational for many de-identification scenarios. In
addition, we make the following observations.

\begin{figure}
    \includegraphics[width=\columnwidth]{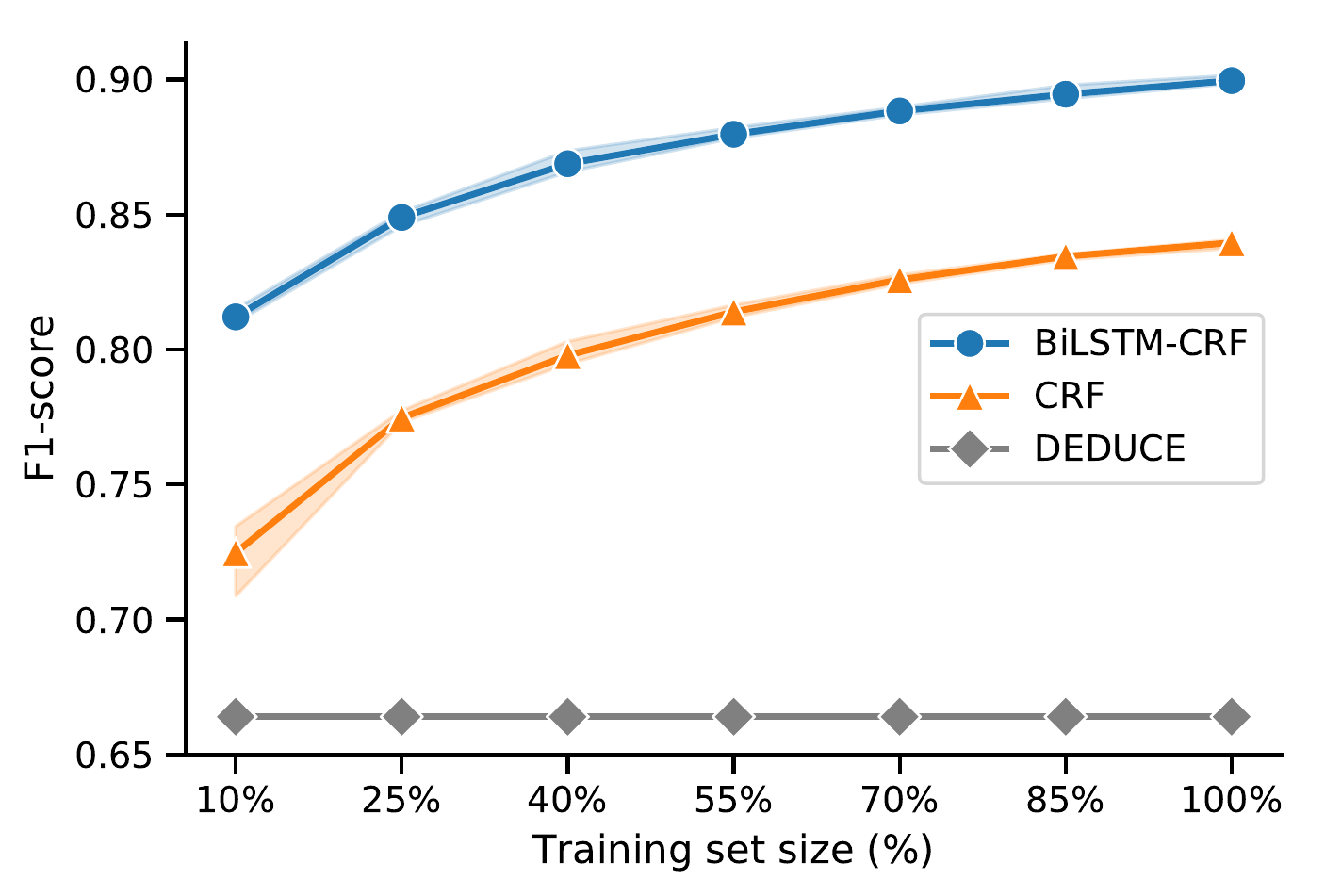}
    \caption{%
    Entity-level F1-score for varying training set sizes. The full training set (100\%) consists of
    all training and validation sentences in NUT (34,714). The F1-score is measured on the test set. For
    each subset size, we draw 3 random samples and train/test each model 3 times. The lines show
    the averaged scores along with the 95\% confidence interval. The rule-based tagger DEDUCE is
    shown as a baseline.
    }
    \label{fig:training-data-vs-performance}
\end{figure}

\textbf{Neural method performs at least as good as rule-based method}.
By inspecting the model performance on a PHI-tag level, we observe that the neural method
outperforms DEDUCE for all classes of PHI (see~\cref{tab:phi-breakdown}).
Only for the \textit{Phone} and \textit{Email} category, the rule-based method has a slightly
higher precision.
Similarly, we studied the impact of the training data set size on the de-identification performance.
Both machine learning methods outperform DEDUCE even with as little training data as 10\% of the
total sentences (see~\cref{fig:training-data-vs-performance}).
This suggests that in most environments where training data are available (or can be obtained), the
machine learning methods are to be preferred.

\textbf{Rule-based method can provide a ``safety net.''}
It can be observed that DEDUCE performs reasonably well for names, phone numbers, email addresses
and URLs (see~\cref{tab:phi-breakdown}).
As these PHI instances are likely to directly reveal the identity of an individual, their removal is essential.
However, DEDUCE does not generalize beyond the PHI types mentioned above.
Especially named locations are non-trivial to capture with a rule-based system as their
identification strongly relies on the availability of exhaustive lookup lists.
In contrast, the neural method provides a significant improvement for named locations (5.8\% vs.
65.9\% recall).
We assume that word-level and character-level embeddings provide an effective tool to capture these
entities.

\textbf{Initials, IDs and professions are hard to detect.}
During annotation, we observed a low F1 annotator agreement of 0.46, 0.43, and 0.31 for initials,
IDs and professions, respectively. This shows that these PHI types are among the hardest to
identify, even for humans (see~\cref{tab:phi-distribution}).
One possible cause for this is that IDs and initials are often hard to discriminate from
abbreviations and medical measurements.
We observe that the BiLSTM-CRF detects those PHI classes with high precision but low recall.
With respect to professions, we find that phrases are often wrongly tagged.
For example, colloquial job descriptions (e.g., ``works behind the cash desk'') as opposed to
the job title (e.g., ``cashier'') make it infeasible to tackle this problem with lookup lists,
while a machine learner likely requires more training data to capture this PHI.

\subsection{Error Analysis on Dutch Dataset}\label{sec:error-analysis}
To gain a better understanding of the best performing model and an intuition for its limitations,
we conduct a manual error analysis of the false positives (FPs) and false negatives (FNs) produced
by the BiLSTM-CRF on the test set. We discuss the error categorization scheme
in~\cref{sec:error-categorization} and present the results
in~\cref{sec:error-analysis-results}.

\subsubsection{Error Categorization}\label{sec:error-categorization}
We distinguish between two error groups: (1) modeling errors, and (2) annotation/preprocessing
errors.
We define modeling errors to be problems that can be addressed with different de-identification
techniques and additional training data.
In contrast, annotation and preprocessing errors are not directly caused by the sequence labeling
model, but are issues in the training data or the preprocessing pipeline which need to be addressed
manually.
Inspired by the classification scheme of~\citet{Dernoncourt:2017:DPN}, we consider the following
sources of modeling errors:
\begin{itemize}
    \setlength{\parskip}{0pt}
    \item \textbf{Abbreviation.} PHI instances which are abbreviations or acronyms for names, care
    institutes and companies. These are hard to detect and can be ambiguous as they are easily
    confused with medical terms and measurements.
    \item \textbf{Ambiguity.} A human reader may be unable to decide whether a given text fragment
    is PHI.
    \item \textbf{Debatable.} It can be argued that the token should not have been annotated as PHI.
    \item \textbf{Prefix.} Names of internal locations, organizations and companies are often
    prefixed with articles (i.e., ``de'' and ``het''). Sometimes, it is unclear whether the prefix
    is part of the official name or part of the sentence construction. This ambiguity is reflected
    in the training data which causes the model to inconsistently include or exclude those prefixes.
    \item \textbf{Common Language.} PHI instances consisting of common language are hard
    to discriminate from the surrounding text.
    \item \textbf{Other.} Remaining modeling errors that do not fall into the categories mentioned
    above. In those cases, it is not immediately apparent why the misclassification occurs.
\end{itemize}
Preprocessing errors are categorized as follows:
\begin{itemize}
    \setlength{\parskip}{0pt}
    \item \textbf{Missing Annotation.} The text fragment is PHI, but was missed during the
    annotation phase.
    \item \textbf{Annotation Error.} The annotator assigned an invalid entity boundary.
    \item \textbf{Tokenization Error.} The annotated text span could not be split into a compatible
    token span. Those tokens were marked as ``Outside (O)'' during BIO tagging.
\end{itemize}
We consider all error categories to be mutually exclusive.

\begin{table}[t]
\caption{Summary of the manual error analysis of false negatives (FNs) and false
positives (FPs) produced by the BiLSTM-CRF. All error categories are mutually exclusive.}
\label{tab:error-analysis-bilstmcrf}
\begin{tabular}{@{}p{.43\columnwidth}rrrr@{}}
\toprule
 & \multicolumn{2}{c}{FNs ($n=469$)} & \multicolumn{2}{c}{FPs ($n=288$)} \\
 \cmidrule(lr){2-3}
 \cmidrule(lr){4-5}
Category & Count & Part & Count & Part \\\midrule
\multicolumn{5}{@{}l}{\emph{Model Errors}}  \\
\hspace{.8em}Abbreviation & 65 & 13.9\% & 28 & 9.7\% \\
\hspace{.8em}Ambiguity & 15 & 3.2\% & 7 & 2.4\% \\
\hspace{.8em}Debatable & 7 & 1.5\% & 4 & 1.4\% \\
\hspace{.8em}Prefix & 10 & 2.1\% & 10 & 3.5\% \\
\hspace{.8em}Common language & 35 & 7.5\% & 9 & 3.1\% \\
\hspace{.8em}Other reason & 275 & 58.6\% & 159 & 55.2\% \\\midrule
\multicolumn{5}{@{}l}{\emph{Annotation/Preprocessing Errors}}  \\
\hspace{.8em}Missing Annotation & - & - & 33 & 11.5\% \\
\hspace{.8em}Annotation Error & 21 & 4.5\% & 18 & 6.3\% \\
\hspace{.8em}Tokenization Error & 41 & 8.7\% & 20 & 6.9\% \\\midrule
Total & 469 & 100\% & 288 & 100\% \\ \bottomrule
\end{tabular}
\end{table}

\subsubsection{Results of Error Analysis}\label{sec:error-analysis-results}
\cref{tab:error-analysis-bilstmcrf} summarizes the error analysis results and shows the absolute
and relative frequency of each error category. Overall, we find that the majority of modeling
errors cannot be easily explained through human inspection
(``Other reason'' in~\cref{tab:error-analysis-bilstmcrf}). The remaining errors are mainly caused
by ambiguous PHI instances and preprocessing errors.
In more detail, we make the following observations:\\
\textbf{Abbreviations are the second most common cause for modeling errors} (13.9\% of FNs, 9.7\% of FPs).
We hypothesize that more training data will likely not in itself help to correctly identify this
type of PHI.
It is conceivable to design custom features (e.g., based on shape, positioning in a sentence,
presence/absence in a medical dictionary) to increase precision.
However, it is an open question how recall can be improved.\\
\textbf{PHI instances consisting of common language are likely to be wrongly tagged} (7.5\% FNs, 3.1\% FPs).
This is caused by the fact that there are insufficient training examples where common language is
used to refer to PHI.
For example, the organization name in the sentence ``Vandaag heb ik \underline{Beter Horen}
gebeld'' (Eng: ``I called \underline{Beter Horen} today'') was incorrectly classified as non-PHI.
Each individual word, and also the combination of the two words, can be used in different contexts
without referring to PHI. However, in this specific context, it is apparent that ``Beter Horen''
must refer to an organization.\\
\textbf{A substantial amount of errors is due to annotation and preprocessing issues.}
Annotation errors (4.5\% FNs, 6.3\% FPs) can be resolved by correcting the respective
PHI offsets in the gold standard. Tokenization errors (8.7\% FNs, 6.9\% FPs) need to be fixed
through a different preprocessing routine.
For example, the annotation \texttt{<DATE 2016>/<DATE 2017>} should have
been split into \texttt{[2016, /, 2017]} with BIO tagging \texttt{[B, O, B]}. However, the spaCy
tokenizer segmented this text into a single token \texttt{[2016/2017]}.
In this case, entity boundaries do no longer align with token boundaries which results in an
invalid BIO tagging of \texttt{[O]} for the entire span.\\
\textbf{Several false positives are in fact PHI and should be annotated.}
The model identifies several PHI instances which were missed during the annotation phase
(11.5\% of the FPs).
Once more, this demonstrates that proper de-identification is an error-prone task for human
annotators.

\subsection{De-identification of English Datasets}\label{sec:results-english-data}
When training and testing both machine learning methods on the English i2b2 and the nursing notes
datasets, we can observe that the BiLSTM-CRF significantly outperforms the CRF in both cases
(see~\cref{tab:results-summary}).
Similar to our Dutch dataset, the neural method provides an increase of up to 11.2\% points in
recall (nursing notes) while the precision remains relatively stable.
This shows that the neural method has the best generalization capabilities even across languages.
More importantly, it does not require the development of domain-specific lookup lists or
sophisticated pattern matching rules.
To put the results into perspective: the second-highest ranked team in the i2b2 2014 challenge used
a sophisticated ensemble combining a CRF with domain-specific rules~\cite{Stubbs:2015:ASD}. Their
system obtained an entity-level F1 score of 0.9124 which is on-par with the performance of our
neural method that requires no configuration.
We can expect that the performance of the neural method further improves after hyperparameter optimization.
Finally, note that both machine learning methods can be easily applied to a new
PHI tagging scheme, whereas rule-based methods are limited to the PHI definition they were
developed for.

\subsection{Cross-domain De-identification}\label{sec:results-cross-domain}
In many de-identification scenarios, heterogeneous training data from multiple medical institutes
and domains are rarely available.
This raises the question, how well a model that has been trained on a homogeneous set of medical
records generalizes to records of other medical domains.
We trained the three de-identification methods on one domain of Dutch healthcare (e.g., elderly
care) and tested each model on the records of the remaining two domains (e.g., disabled care and
mental care).
We followed the same training and evaluation procedures described in~\cref{sec:evaluation}.
~\cref{tab:transfer-learning-summary} summarizes the performance of each method on the
different tasks.

Again, the neural method consistently outperforms the rule-based and feature-based methods in all
three domains which suggests that it is a fair default choice for de-identification.
This is underlined by the fact that the amount of training data is severely limited in this
experiment: each domain only has 420 documents of which 20\% of the records are reserved for
testing.
Interestingly, DEDUCE performs rather stable and even outperforms the CRF within the domain of
elderly care.

\begin{table}[t]
\caption{Summary of the transfer learning experiment on our Dutch dataset. Each method is trained
on data of one care domain and tested on the other two domains. All scores are micro-averaged
entity-level F1.}
\label{tab:transfer-learning-summary}
\begin{tabular}{@{}lrrr@{}}
\toprule
 & \multicolumn{3}{c}{Training Domain}  \\ \cmidrule(lr){2-4}
Method & Elderly & Disabled & Mental \\ \midrule
DEDUCE & 0.683 & 0.565 & 0.675 \\
CRF & 0.414 & 0.697 & 0.719 \\
BiLSTM-CRF & \textbf{0.775} & \textbf{0.775} & \textbf{0.839} \\ \bottomrule
\end{tabular}
\end{table}

\begin{table}[t]
\caption{Detailed performance analysis of the BiLSTM-CRF method in the transfer learning
experiment. In-domain test scores are shown on the diagonal.
All scores are micro-averaged entity-level F1.}
\label{tab:transfer-learning-bilstmcrf}
\begin{tabular}{@{}lrrr@{}}
\toprule
  & \multicolumn{3}{c}{Training Domain}  \\ \cmidrule(lr){2-4}
Test Domain & Elderly & Disabled & Mental \\ \midrule
Elderly & 0.746 & 0.698 & 0.703 \\
Disabled & 0.796 & 0.919 & 0.879 \\
Mental & 0.744 & 0.806 & 0.871 \\ \bottomrule
\end{tabular}
\end{table}

Given an ideal de-identification method, one would expect that performance on unseen data of a
different domain is similar to the test score obtained on the available (homogeneous) data.
\cref{tab:transfer-learning-bilstmcrf} shows a performance breakdown for each of the three testing
domains for the neural method.
It can be seen that in 4 out of 6 cases, the test score in a new domain is lower than the test
score obtained on the in-domain data.
The largest delta of the observed in-domain test score (disabled care, 0.919 F1) and the
performance in the transfer domain (elderly care, 0.698 F1) is 0.221 in F1. This raises an
important point when performing de-identification in practice:
while the neural method shows the best generalization capabilities compared to the other
de-identification methods, the performance can still be significantly lower when applying a
pre-trained model in new domains.

\subsection{Limitations}
While the contextual string embeddings used in this paper have shown to provide state-of-the-art results for NER~\cite{Akbik:2018:CSE}, transformer-based architectures for contextual embeddings have also gained significant attention (e.g., BERT~\cite{Devlin:2018:BPD}).
It would make an interesting experiment to benchmark different types of pre-trained embeddings for the task of de-identification.
Furthermore, we observe that the neural method provides strong performance even with limited training data (see~\cref{fig:training-data-vs-performance}).
It is unclear what contribution large pre-trained embeddings have in those scenarios which warrants an ablation study testing different model configurations.
We leave the exploration of those ideas to future research.

%% file: 90-conclusion.tex
\section{Conclusion}\label{sec:conclusion}
This paper presents the construction of a novel Dutch dataset and a comparison of state-of-the-art
de-identification methods across Dutch and English medical records.
Our experiments show the following.
(1) An existing rule-based method for the Dutch language does not generalize well to new domains.
(2) If one is looking for an out-of-the-box de-identification method, neural approaches show the
best generalization performance across languages and domains.
(3) When testing across different domains, a substantial decrease of performance has to be
expected, an important consideration when applying de-identification in practice.

There are several directions for future work. Motivated by the limited generalizability of
pre-trained models across different domains, transfer learning techniques can provide a way
forward.
A preliminary study by~\citet{Lee:2018:TLN} shows that they can be beneficial for
de-identification.
Finally, our experiments show that phrases such as professions are among the most difficult
information to de-identify.
It is an open challenge how to design methods that can capture this type of information.